\documentclass[11pt]{article}
\usepackage{acl2012}
\usepackage{times}
\usepackage{latexsym}
\usepackage{amsmath}
\usepackage{multirow}
\usepackage{url}
\usepackage{graphicx}
\usepackage{tikz-dependency}
\usepackage{verbatim}

\title{Classifying Syntactic Regularities for Hundreds of Languages}

\author{Reed Coke \\
  Department of EECS \\
  University of Michigan \\
  Ann Arbor \\
  {\tt reedcoke@umich.edu} \\\And
  Ben King \\
   Ubiquiti, Inc.\\
  {\tt benking@ubiquiti.com} \\\And
  Dragomir Radev \\
  Department of EECS \\
  School of Information \\
  University of Michigan \\
  Ann Arbor \\
  {\tt radev@umich.edu} \\}

\date{}

\begin{document}
\maketitle
\begin{abstract}
This paper presents a comparison of classification methods for
linguistic typology for the purpose of expanding an extensive, but sparse
language resource: the World Atlas of Language Structures (WALS) \cite{wals}.
We experimented with a variety of regression and
nearest-neighbor methods for use in classification over a set of 325
languages and six syntactic rules drawn from WALS. To classify each rule, we
consider the typological features of the other five rules; linguistic
features extracted from a word-aligned Bible in each language; and
genealogical features (genus and family) of each language. In general,
we find that propagating the majority label among all languages of the
same genus achieves the best accuracy in label prediction. Following
this, a logistic regression model that combines typological and
linguistic features offers the next best performance. Interestingly, this
model actually outperforms the majority labels among all languages of
the same family.

\end{abstract}

\section{Introduction}
\label{intro}

Linguistic typology is a subfield of linguistics concerned with
understanding the various patterns that are present across the world's
languages and how languages can be grouped via these patterns as well
as other historic and geographic factors. Other fields of linguistics,
such as historical linguistics and the study of endangered languages,
depend heavily on knowledge drawn from typological
comparisons. Linguistic typology is also useful in many areas of
natural language processing. 

WALS is an expansive linguistic resource useful in a variety of
different NLP applications. WALS has data for over 2,500 languages
regarding almost 200 rules compiled by a group of 55 authors
spanning phonology, syntax, and lexicology. Examples of such rules and their
possible values are given in section \ref{darules}. WALS has been used
as a lens through which to view worldwide typological relations
\cite{Littauer12} as well as a tool with which to discover which
linguistic rules are rare across the world's languages
\cite{Cysouw11}. Unsurprisingly, it has also served as a baseline for
typological similarity measurements \cite{Berzak14}.
In this work the authors also show that the accuracy of the
baseline increases with the number of WALS rules considered,
but note the sparseness of these rules.
Indeed, the sparseness of WALS has been noted by
several researchers \cite{Georgi10,Cysouw08,Teh08}. In fact, currently
the average rule in WALS has data for only 400 languages. This
represents a matrix that is only 16\% populated. To this end, we
examine methods for automatically identifying rule values in the WALS
database for the purpose of reducing sparseness in WALS. Our methods
use the WALS database as of March 2015 as well as an additional copy of the
Bible in each target language downloaded from \url{bible.com} or \url{bible.is}
and word-aligned to an English version. 

Despite its sparseness, WALS is still an excellent resource. Consider
a similar resource, Syntactic Structures of the World's Languages
(SSWL) \cite{sswl}. SSWL is considerably less sparse, with 56\% of all
possible data for 112 rules and 251 languages. In reality, many of
these rules actually represent pieces of a single rule in WALS - SSWL
rule 3 is \emph{Verb Object} while SSWL rule 4 is \emph{Object
  Verb}. In WALS, both of these rules are possible values for rule 83A:
Order of Object and Verb. Thus, SSWL lacks the breadth that makes WALS
such an appealing resource. WALS has an order of magnitude more languages
and many more rules. In addition, WALS has genealogical information about
each language that SSWL lacks.

While many of the rules in WALS, such as rule 83A: Order of Object and
Verb, may seem simple, properties of linguistic universals can be
exploited to extrapolate further information. Joseph Greenberg
proposed a set of linguistic universals \cite{Greenberg63} that refer
to statistical tendencies across the world's languages. Further
universals have been described since, particularly with respect to
word order \cite{Hawkins83}. While our system does not directly make
use of these universals, the core mechanic of classifying rules based
on the labels of other rules for the same language is central to our
experimentation. This is the grounding theory behind the use of
typological features.

Applying effective, automated methods for expanding this database, or
similar databases, will not only create a better resource, but will
also lead to improvements in multilingual NLP via better parsing, a
better understanding of linguistic typology, and many other basic NLP
tasks such as machine translation \cite{Mikolov13} and part-of-speech
tagging \cite{Das11}. In addition, having a more expansive resource
will allow for NLP to be leveraged in many under-resourced languages
not only in Europe, but around the world.

The rest of this paper is structured as follows: Section
\ref{relatedWork} discusses other work in the field of NLP that has
made of use of features in WALS; Section \ref{exps} describes in
detail how we generated our feature vectors and present the
classification models we compared; Section \ref{RandD} presents our
results and provides an analysis; Section \ref{conc} concludes our
work; and Section \ref{future} suggests future work. 

\section{Related Work}
\label{relatedWork}

Typological similarity has previously been shown to correlate with
genealogical similarity both in the fields of NLP \cite{Rama12} and
historical linguistics \cite{Dunn05}. Due to this, WALS has been used
to study linguistic typology via computational methods. In order to
determine language similarity via genealogical relatedness, it is
important to know which rules are more telling of historical
relatedness. To do this, \cite{Wichmann08} measured the variance of
linguistic rules within language families, at the genus level, to
determine their stability. Rules that change less often within
language families, then, are more indicative of historical
relationships when they are shared between languages. To do this, the
authors extracted data for language families from WALS and calculated
the probability of features being shared within and across language
families. Feature variation was shown to be significantly different
across genera. This indicates that language families are more similar
internally than across genera, which in turn supports the concept of
using typological similarity to predict language similarity. 

These syntactic regularities are not only predictable within language
genera, however.  \cite{DaumeImpls} used WALS as a database to
discover implicative associations, similar to Greenberg's
universals. Not only did they recover many of Greenberg's universals,
but they also uncovered a host of other implicatures.

In similar work, \cite{Cysouw08} sought to determine how consistent
certain linguistic rules are in an effort to determine languages that
are ``more central for the structure of human language."
The authors cite ``widely varying frequencies of available data" as
a difficulty in their study, but conclude that many word order features
seem to be central rules in language. Among these are rules 83A,
85A, 86A, 88A, and 107A, five of the six rules considered in this study. 

Apart from typology, the data from WALS have also been used in natural
language processing to advance several core NLP tasks. \cite{Naseem12}
used language similarity information to improve multilingual parsing
by defined a distance metric over WALS rules. The authors considered
six WALS rules to create their similarity metric, three of which
(85A, 86A, and 88A), are also used in this study. 

\cite{Bender13} attempted a similar task, using interlinear glossed
text (IGT) to predict typological features such as major consituent
order and case system.  \cite{Lewis2008} also predict typological
features from IGT by first learning a context free grammar for a
language and then examining its structure. Both studies use data
from WALS to evaluate their performance.  We address a similar task,
but we utilize projected dependency parses from English text as a
source of knowledge for each foreign language, as many do not
have extensive resources of their own. This is similar to IGT, but
does not contain morphological information to the same extent. This
method of projecting dependencies has been used to bootstrap
linguistic resources in the past \cite{Xia07}, \cite{Hwa05}.

\section{Experiments}
\label{exps}

We first created a corpus of Bibles word-aligned to an English
Bible. The 325 languages represented in the corpus are very diverse,
ranging from French, German, and Modern Standard Arabic to Acholi,
Basque, and Tamil. Note that translated Bibles exist for many other languages
that were not included in this study. In total, we use over 2 million
aligned sentences. The rules were selected based on the frequency of
their use in previous NLP research, a measure of usefulness. More
information about them is provided in Section \ref{darules}. For each
rule we ran experiments using a variety of classifiers and feature
vector combinations. All results are given in Section \ref{RandD}.

\subsection{General Textual Feature Extraction}

For each rule described in Section \ref{darules}, we extract English
dependencies according to certain criteria unique to that rule. We
then project these dependencies onto the foreign language biblical
sentences using word alignments determined by BerkeleyAligner
\cite{DeNero07,Liang06}. Using these inferred dependencies, we
calculate the feature vectors for the rule in question by the policy
described in Section \ref{darules}. 

\begin{figure}
\centering
\scalebox{0.4}{
\includegraphics[width=\textwidth]{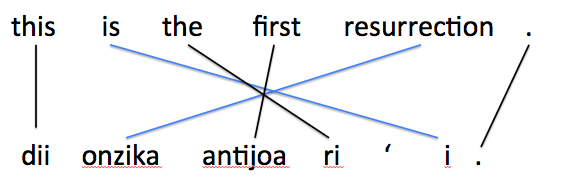}
}
\caption{Word alignments for an example sentence in Ma'di and English}
\label{madiEx}
\end{figure}

To give a simple example of projecting word alignments,
consider the English sentence and its Ma'di translation in
Figure \ref{madiEx}. \emph{is} and \emph{resurrection}
are the verb and object of the English sentence, while \emph{i} and
\emph{onzika} are the Ma'di alignments. These alignments are shown
using blue edges. As these words are aligned, we assume they serve the
same grammatical function. We can see that the English sentence shows
a verb-object ordering, while the Ma'di sentence demonstrates an
object-verb ordering. This would then add one to the count of
object-verb orderings to the linguistic feature vector of Ma'di for
rule 83A. 

\begin{figure*}
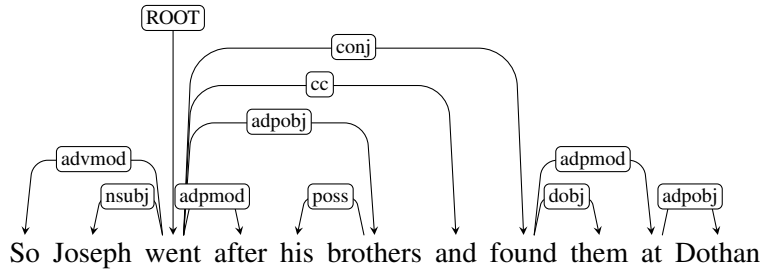

\centering
\scalebox{1}{
\begin{dependency}
\begin{deptext}
So \& Joseph \& went \& after \& his \& brothers \& and \& found \& them \& at \& Dothan \\
\end{deptext}
\deproot[edge unit distance=4.5ex]{3}{ROOT}
\depedge{3}{1}{advmod}
\depedge{3}{2}{nsubj}
\depedge{6}{5}{poss}
\depedge{3}{6}{adpobj}
\depedge{3}{7}{cc}
\depedge{3}{8}{conj}
\depedge{8}{9}{dobj}
\depedge{3}{4}{adpmod}
\depedge{8}{10}{adpmod}
\depedge{10}{11}{adpobj}
\end{dependency}
}
\caption{Sample English dependency parse}
\label{englishParse}
\end{figure*}

\begin{figure*}
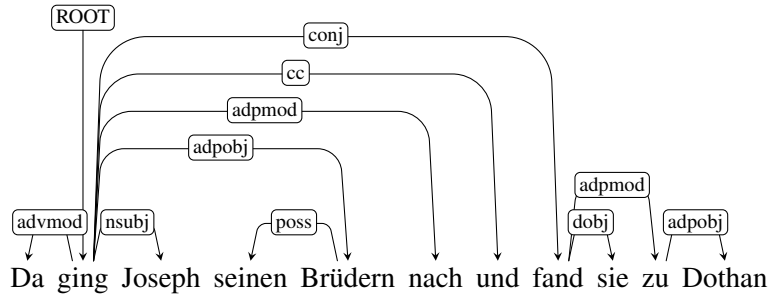

\centering
\scalebox{1}{
\begin{dependency}
\begin{deptext}
Da \& ging \& Joseph \& seinen \& Br\"{u}dern \& nach \& und \& fand \& sie \& zu \& Dothan \\
\end{deptext}
\deproot[edge unit distance=5ex]{2}{ROOT}
\depedge{2}{1}{advmod}
\depedge{2}{3}{nsubj}
\depedge{2}{6}{adpmod}
\depedge{5}{4}{poss}
\depedge{2}{5}{adpobj}
\depedge{2}{7}{cc}
\depedge{2}{8}{conj}
\depedge{8}{10}{adpmod}
\depedge{10}{11}{adpobj}
\depedge{8}{9}{dobj}
\end{dependency}
}
\caption{Sample German dependency parse}
\label{germanParse}
\end{figure*}

Next, we give a detailed example of how the features for WALS rule
83A \emph{Order of Object and Verb} can be drawn from projected
dependency parses of English text as in the German example in
Figures \ref{englishParse} and \ref{germanParse}. 

In this example, German would receive one count for having the same
verb-object ordering in any phrase due to the (\emph{Dothan},
\emph{found}) dependency and one count for different ordering due to
the (\emph{brothers}, \emph{after}) dependency. This sentence is not a
question or a dependent clause, so the corresponding same/different
counts would not be changed for those features. This process is then
repeated for all aligned sentences in all languages with a known label
for the rule. 

\subsection{Specific Feature Extraction}
\label{darules}
Each rule is described in detail below, along with definitions of its
possible classes, rounded distributional information over the
languages considered in this paper, and a brief motivation for why we
chose to classify this rule. All parses referred to in this section
were obtained using a Hash Kernel parser \cite{bohnet10} and
represented in the CONLL dependency parsing format. We restrict our
system to instances where the English sentence aligns to only one
foreign language sentence. The count of languages given for each rule
is the size of the set of languages whose label for this rule is given
in WALS and for whom we have a Bible. 

\subsubsection{Rule 83A: Order of Object and Verb}
Rule 83A defines the order in which the direct object and verb of a
sentence occur. Its label is known for 299 languages. The three
possible classes are \emph{Object Verb} (The man saw the dog. 38\%),
\emph{Verb Object} (The man the dog saw. 58\%), and \emph{No Dominant
  Order} (both valid in certain conditions. 4\%). Besides being
fundamental to any sort of parsing or chunking task, many other
typological rules have been found to be statistically related to the
order of subject, object, and verb \cite{Greenberg63}. As it has also
been found that subject generally precedes object \cite{Greenberg63},
knowing the order of the object and verb allows not only accurate
guesses about the order of subject, object, and verb, but also about
many other correlated rules. To identify all occurrences, we examine
all dependencies with the label \emph{dobj}. The instance is kept as
long as both of the English and foreign object words align only to
each other and both of the English and foreign verb words align only
to each other. The resulting feature vector has six columns: the count
of all sentences in which the verb-object ordering is identical
between English and the foreign language; the count of all sentence in
which the ordering is reversed; the count of all questions in which
the ordering is identical; the count of all questions in which it is
reversed; and finally the count of dependent clauses where the order
is identical and the count of dependent clauses in which the order is
reversed.

\subsubsection{Rule 85A: Order of Adposition and Noun Phrase}
Rule 85A defines the order in which adpositions occur relative to
their governing noun phrase. Its label is known for 256 languages. The
five possible classes are \emph{postpositions} (The dog went the park
to. 43\%), \emph{prepositions} (The dog went to the park. 49\%),
\emph{inpositions} (The dog went the to park. 0\%), \emph{more than
  one type with none dominant} (multiple valid. 6\%), and \emph{no
  adpositions} (The dog went the park. 1\%). This rule is critical for
having any sort of successful semantic parse or narrative
processing. We examine all dependencies with the label \emph{adpmod}
and have a parent word in the sentence. The instance is kept as long
as the English and foreign adpositions align only to each other and
the English and foreign governing nouns align only to each other. The
feature vector has a dimensionality of six and consists of the same
features as the feature vector for Rule 83A, simply computed for the
order of the adposition and noun phrase, rather than object and verb.

\subsubsection{Rule 86A: Order of Genitive and Noun}
Rule 86A defines the order in which genitives appear with respect to
their governing noun. 254 languages in our dataset have a known label
for this rule. The three possible classes are \emph{Genitive-noun} (My
dog 47\%), \emph{Noun-genitive} (Dog my 45\%), and \emph{Both occur,
  none dominant} (Both valid 8\%). This rule is absolutely necessary
for coreference resolution. We examine all dependencies with the label
\emph{poss}, as long as the genitive and noun align only to their
respective foreign language counterpart. Again, we use a length six
feature vector with the same procedure over this set of dependencies. 

\subsubsection{Rule 88A: Order of Demonstrative and Noun}

Rule 88A defines the order in which demonstratives appear together
with nouns. There are 171 languages in our set of Bibles with a known
label for this rule. The six possible classes are \emph{Demonstrative
  noun} (This dog is shaggy. 50\%), \emph{Noun demonstrative} (Dog
this is shaggy. 42\%), \emph{prefix on noun} (This-dog is
shaggy. 0\%), \emph{suffix on noun} (Dog-this is shaggy. 2\%),
\emph{Demonstrative noun demonstrative} (This dog this is
shaggy. 2\%), and \emph{two or more of the previous options, none
  dominant} (multiple valid. 4\%). This rule is also necessary for
coreference resolution. We select all dependencies that are of type
\emph{det} or \emph{pron}, have a lexical value of \emph{this},
\emph{that}, \emph{these}, or \emph{those}, and whose parent is a
noun. We compute the feature vector in the same fashion as the above
rules. 

\subsubsection{Rule 92A: Position of Polar Question Particles}

Rule 92A defines the position that a polar question particle appears
in a sentence. Its rule is known for 91 languages. Polar question
particles signal grammatically that the sentence is a yes or no
question. The six possible classes are \emph{beginning of sentence}
(23\%), \emph{end of sentence} (36\%), \emph{second word in sentence}
(3\%), \emph{anywhere else} (2\%), \emph{either of two positions}
(1\%), \emph{no question particle} (34\%). In languages with polar
questions, this rule would allow for much higher quality question
answering and dialogue systems. Our selection of this rule also
demonstrates that the phenomenon does not even need to occur in the
source language in order to be approached. We select all dependencies
that contain questions. We then use the presence of an English wh-
word to sort these questions into polar and information
questions. Using these labels, we examine the foreign language
questions for the word that appears most often in polar questions but
not information questions using the ratio of relative frequencies. We
then count the number of times this word appears in each of the
possible positions. The feature vector has a length of four, and each
entry corresponds directly to a label class: we count how often the
inferred question word appears initially, second, last, or elsewhere. 

\subsubsection{Rule 107A: Passive Constructions}

107A describes the presence or absence of passive constructions. We have
Bibles for 93 of the WALS languages with a known label for this
rule. The two possible classes are \emph{passive construction} (The
dog was seen by the butler. 53\%) and \emph{no passive construction}
(The butler saw the dog. 47\%). This rule was chosen not only because
it would allow for better grammar induction, a very relevant task for
such under-resourced languages, but also to demonstrate that our
system and approach can handle rules other than simple word order
rules. We select all dependencies that are of type \emph{nsubjpass},
signaling a passive subject, and \emph{nsubj}, signaling an active
subject. We then determine the number of times that the order of
subject and verb differs between these two sentence types in the
foreign language. The feature vector is built identically to all other
rules except for 92A. 

\subsection{Text Classifier Training}

In order to train the text classifiers, we used word-aligned biblical
texts in English and each of the languages considered along with the
labels from WALS. For each rule discussed in Section \ref{darules}, we
select dependencies and create normalized feature vectors according to
the discussed method. This forms the Text feature set in Section
\ref{RandD}. 

\subsection{Typological Classifier Training}

We also consider that some labels can be predicted from the labels of
other rules within a language. We create feature vectors for each
language using five of the six rules discussed in Section
\ref{darules} in order to classify the sixth rule. This comprises the
Rules feature set in Section \ref{RandD}. 

\subsection{Genealogical Classifier Training}

Previous work has shown that propagating knowledge from genealogically
similar languages has demonstrated markedly better results than
propagating knowledge from a random language \cite{Naseem12}. For our
purposes we consider the genus and family of each language as given by
WALS. To this end, we consider simply propagating the majority label
from all languages of the same genus and, separately, all languages of
the same family.  We also create feature vectors such as those for the
Text and Rules features, as described above.

\section{Results and Discussion}
\label{RandD}

\emph{Majority} is simply the majority class as shown in
\ref{darules}. Na\"{i}ve Bayes is run using the default implement in
Weka \cite{Hall09}. For logistic regression, as we do not have a
developmental data set, we simply experiment with five common
regularization parameters - 1.0, 0.5, 0.1, 0.01, and $10^{-8}$ (the
default value in Weka). As we are only evaluating over languages that
already have known labels in WALS, we consider the accuracy of each
classifier through leave-one-out cross-validation.

\subsection{Simple Linguistic Features}

\begin{table*}[h!tb]
\centering
\scalebox{1}{
\begin{tabular}{|c|c|c|c|c|c|c|c|}
\hline
Features & Majority & Na\"{i}ve Bayes & LR1 & LR.5 & LR.1 & LR.01 & LR-8 \\
\hline
Text & 58.0\% & 92.0\% & \textbf{92.3\%} & \textbf{92.3\%} & \textbf{92.3\%} & \textbf{92.3\%} & 92.0\%\\
\hline
Rules & 58.0\% & 82.2\% & 83.3\% & 83.3\% & 83.6\% & 83.6\% & 83.6\% \\
\hline
Text+Rules & 58.0\% & 90.6\% & 90.6\% & 90.0\% & 88.9\% & 89.6\% & 89.0\% \\
\hline
\end{tabular}}
\caption{Accuracy for rule 83A: Order of Object and Verb N=299}
\label{results83A}
\end{table*}

It is clear from Tables \ref{results83A} - \ref{results107A} that the
combination of textual and typological features works best. It can
also be seen that almost regardless of the regularization parameter,
logistic regression models this data with higher accuracy than
na\"{i}ve Bayes. In the cases that this is not true, rule 83A and rule
88A, it is the linguistic features that perform best. This is also
encouraging, as the main problem we are trying to address is the
sparseness of the WALS data. 

There are also two rules that perform quite poorly, rule 92A and rule
107A, depicted in Tables \ref{results92A} and \ref{results107A}. These
rules are significantly more subtle linguistically than the rest,
which are relatively straightforward for our parser to detect
accurately. The complexity of the pipeline for rule 92A and the
numerous ways in which languages can mark passive constructions,
regarding rule 107A, are simply difficult to perceive. 

\begin{table*}[h!tb]
\centering
\scalebox{1}{
\begin{tabular}{|c|c|c|c|c|c|c|c|}
\hline
Features & Majority & Na\"{i}ve Bayes & LR1 & LR.5 & LR.1 & LR.01 & LR-8 \\
\hline
Text & 49.6\% & 82.0\% & 83.2\% & 83.2\% & 84.4\% & 84.4\% & 84.0\%\\
\hline
Rules & 49.6\% & 82.0\% & 84.8\% & 84.8\% & 84.4\% & 84.0\% & 83.6\%\\
\hline
Text+Rules & 49.6\% & 83.2\% & \textbf{86.7\%} & 85.5\% & 84.8 & 83.2\% & 82.4\%\\
\hline
\end{tabular}}
\caption{Accuracy for rule 85A: Order of Adposition and Noun Phrase N=256}
\label{results85A}
\end{table*}

\begin{table*}[h!tb]
\centering
\scalebox{1}{
\begin{tabular}{|c|c|c|c|c|c|c|c|}
\hline
Features & Majority & Na\"{i}ve Bayes & LR1 & LR.5 & LR.1 & LR.01 & LR-8 \\
\hline
Text & 47.2\% & 66.9\% & 66.5\% & 66.9\% & 66.1\% & 66.1\% & 66.1\%\\
\hline
Rules & 47.2\% & 81.6\% & 81.3\% & 80.5\% & 80.5\% & 80.1\% & 79.8\%\\
\hline
Text+Rules & 47.2\% & 79.1\% & 81.1\% & 81.5\% & \textbf{81.9\%} & 81.5\% & 81.1\% \\
\hline
\end{tabular}}
\caption{Accuracy for rule 86A: Order of Genitive and Noun N=254}
\label{results86A}
\end{table*}

\begin{table*}[h!tb]
\centering
\scalebox{1}{
\begin{tabular}{|c|c|c|c|c|c|c|c|}
\hline
Features & Majority & Na\"{i}ve Bayes & LR1 & LR.5 & LR.1 & LR.01 & LR-8 \\
\hline
Text & 50.2\% & 74.3\% & 83.6\% & 83.6\% & \textbf{84.2\%} & 84.2\% & 84.2\%\\
\hline
Rules & 50.2\% & 60.8\% & 67.3\% & 66.7\% & 64.9\% & 64.3\% & 63.7\%\\
\hline
Text+Rules & 50.2\% & 74.2\% & 81.3\% & 81.3\% & 80.7\% & 78.9\% & 76.0\% \\
\hline
\end{tabular}}
\caption{Accuracy for rule 88A: Order of Demonstrative and Noun N=171}
\label{results88A}
\end{table*}

\begin{table*}[h!tb]
\centering
\scalebox{1}{
\begin{tabular}{|c|c|c|c|c|c|c|c|}
\hline
Features & Majority & Na\"{i}ve Bayes & LR1 & LR.5 & LR.1 & LR.01 & LR-8 \\
\hline
Text & 36.2\% & 26.4\% & 27.5\% & 27.5\% & 29.7\% & 28.6\% & 27.5\%\\
\hline
Rules & 36.2\% & 38.5\% & 37.4\% & 36.3\% & 40.7\% & 39.6\% & 38.5\%\\
\hline
Text+Rules & 36.2\% & 33.0\% & \textbf{42.9\%} & 35.2\% & 34.1\% & 33.0\% & 31.9\%\\
\hline
\end{tabular}}
\caption{Accuracy for rule 92A: Position of Polar Question Particles N=91}
\label{results92A}
\end{table*}

\begin{table*}[h!tb]
\centering
\scalebox{1}{
\begin{tabular}{|c|c|c|c|c|c|c|c|}
\hline
Features & Majority & Na\"{i}ve Bayes & LR1 & LR.5 & LR.1 & LR.01 & LR-8 \\
\hline
Text & 52.6\% & 50.5\% & 47.3\% & 48.4\% & 47.3\% & 45.2\%& 44.1\% \\
\hline
Rules & 52.6\% & 55.9\% & 55.9\% & 58.1\% & 58.1\% & 58.1\% & 55.9\%\\
\hline
Text+Rules & 52.6\% & 57.0\% & 58.1\% & \textbf{59.1\%} & 57.0 & 55.9\% & 54.8\%\\
\hline
\end{tabular}}
\caption{Accuracy for rule 107A: Passive Constructions N=93}
\label{results107A}
\end{table*}

We also find that regularization does make a difference for the
various logistic regression models.  In some cases, such as Table
\ref{results92A}, the difference can be as large as 10\% accuracy.  In
others, such as Table \ref{results86A}, the difference is only
0.8\%. In general, it appears that a regularization parameter between
1 and 0.5 yields the best, most consistent results for the combined
feature vector case. This is likely due to the nature of the two
feature vectors being combined, as one is inherently discrete while
the other is continuous, but more experimentation is necessary to
determine the relationship more precisely.

\subsection{Genealogy-based Classification}

The genealogical experiments led to rather different results. We first
tested a very simple approach, assigning the majority label for
languages of the same genus or same family. Table \ref{simpleGene}
demonstrates these results.

\begin{table*}[h!tb]
\centering
\scalebox{1}{
\begin{tabular}{|c|c|c|c|}
\hline
Rule & Best Other & Same Genus & Same Family\\
\hline
83A & 92.3\% & \textbf{95.8\%} & 86.3\% \\
\hline
85A & 86.7\% & \textbf{94.3\%} & 84.0\%\\
\hline
86A & 81.9\% & \textbf{93.4\%} & 80.8\%\\
\hline
88A & 84.2\% & \textbf{88.0\%} & 81.3\%\\
\hline
92A & 42.9\% & \textbf{91.1\%} & 73.3\% \\
\hline
107A & 59.1\% & \textbf{95.9\%} & 81.3\% \\
\hline
\end{tabular}}
\caption{Accuracy using Genealogical Neighbors}
\label{simpleGene}
\end{table*}

It is very clear that this majority voting scheme works much better
than the other approaches. This is particularly interesting, given the
argument that typological features have been claimed to be an
effective proxy for genealogical data \cite{Rama12}.  

We also attempted to incorporate the genealogical features as a third
feature source, similar to our textual and typological features.
Combining genealogical features with the either Text or Rule features
did not outperform the systems already presented, nor did the
combination of all three for the na\"{i}ve Bayes or any logistic
regression model. We also considered that perhaps there is something
intrinsic to majority voting that is boosting the performance of this
simple approach. To confirm, we ran experiments on the Text and Rule
feature vectors using $k$-nearest neighbor classification. Again,
these experiments did not rival the performance of our other
classifiers, regardless of the choice of $k$.

It has been estimated that the accuracy of the current WALS database
is possibly close to 96\%, based on examination of the entries for
Latvian \cite{Cysouw11}. The results in Table \ref{simpleGene} are
just shy of this number for many of the rules. This means that for
many rules, the simple genealogical majority measure could be used as
is to expand WALS. Obviously this system will not be as accurate as a
trained linguist, and we are not suggesting that if we expanded WALS
this way it would be fit to replace the current database, but rather
that the system could be used to create a more expansive database at
the cost of slightly lower accuracy.

\section{Conclusion}
\label{conc}

In this study, we examined the effectiveness of classifying a series
of syntactic rules across a set of foreign languages. This is doable
by considering the values of other syntactic rules in the
database. However, it is also worth considering actual linguistic
features for this linguistic task, as well as combining the two
disparate sources. Finally, we also considered genealogical data
available from the WALS database. 

We have shown that with proper regularization, we can achieve better
classification accuracy by augmenting the currently available WALS
data with linguistic features. Though in some cases the individual
linguistic features actually outperform the combined features, this is
not generally the case and it only applies when the linguistic
features are performing well on their own. In addition, it is not
possible to say from our results which regularization parameter is
best. We do, however, demonstrate that this combination of features
can lead to greater success than either feature independently.  This
result is important despite performing worse than the genealogical
approach as the goal of our study is ultimately to find accurate
methods to classify features for languages that are not already
contained in WALS as well as features that are missing for languages
already present.

Considering only the genus of the language, we are within 2\% of
achieving human-like accuracy for half of our six rules. This is
encouraging and further experimentation needs to be done over the
entire set of rules in WALS. However, this approach is limited as the
fundamental problem of sparsity applies to typological data, which is
also used to determine the true testing labels in this case. For this
reason, it is also important that we have demonstrated the usefulness
of purely linguistic features that are separate from the WALS
database, as well as their compatibility with the typological values
in WALS. 

\section{Future Work}
\label{future}

Moving ahead, more complex WALS rules could be addressed with this
method. While the rules we consider are all useful in NLP research,
there are other rules that are almost as frequently used that are more
complex, such as those governing negation. In particular, modules
concerning morphological features that are not easily detectable using
word alignments are the next logical direction. It would also be worth
gathering word aligned Bibles from other well-supported source
languages besides English. As was demonstrated by Rule 92A: Position
of Polar Question Particle, the system can work even when the feature
does not appear in the source language, but it is certainly more
difficult to account for. Having several source languages would
minimize the number of rules for which this is the case.

In addition, experiments with combinations of genealogical, textual,
and typological features should be performed. However, rather than
simply concatenating these features, the features themselves should be
combined into compound features, such as "Germanic and
83A-No\_Dominant\_Order". We believe these types of features could
improve upon the performance of the majority-vote features that are
currently the best measure. The system might then achieve even higher
accuracy, possibly on par with human performance for some rules,
allowing the WALS database to be automatically expanded, perhaps as a
parallel resource to the current database. 

In addition, having shown that this is a reasonably accurate method
for determining values of WALS rules, the method should be applied on
as of yet unlabeled languages in the WALS database.  We also have
nearly 400 such Bibles in languages not contained in WALS, which have
not been used in this experiment. Adding them to WALS would
expand the number of languages in the database by more than ten
percent. While we would not be able to apply the most accurate,
genealogical method, we could certainly make use of linguistic
features from the Bibles. 

It would also be possible to use semi-supervised methods such as
semi-supervised k-Nearest Neighbors and graph regularization
\cite{zhu03} as an improvement to our classification method. Though we
did experiment with k-Nearest Neighbors in this study, it is possible
that a different approach could still yield good results. The idea of
nearest-neighbors classification stands to be very useful, as shown by
the majority label propagation. Graph regularization has been shown to
be effective in weakly-supervised situations where there is sparse
training data \cite{Hassan14}. Therefore, it would be a good idea to
apply it to the WALS data. One could build a similarity network of all
the languages using the known WALS rules that they share in common and
regularize this graph in order to propagate labels to some extent
whenever possible. Constraints such as the genus-majority relationship
discussed in this paper as well as constraints borrowed from
linguistic universals, representing conditional dependencies between
rules, could also be used to help regularize the entire network.

\bibliography{wals}{}
\bibliographystyle{acl2012}

\end{document}